%% file: paper.tex
\documentclass[journal]{IEEEtran}
\usepackage{comment}

\ifCLASSINFOpdf
   \usepackage[pdftex]{graphicx}

\else

\fi

\usepackage{url}
\usepackage[]{algorithm2e}

\usepackage{longtable}
\usepackage{booktabs}
\usepackage{graphicx}

\usepackage{tikz,pgfplots,pgfplotstable}
\usepackage{pgfgantt}
\usepackage{mathtools}
\usepackage{hyperref}
\usetikzlibrary{shapes,intersections}

\tikzset{place/.style = {circle, draw=blue!50, fill=blue!20, thick, minimum size=0.6cm},
    transition/.style = {rectangle, draw, fill=blue!20, stroke=black, thick, minimum width=3cm,
                        minimum height = 3cm},
    pre/.style =    {<-, semithick},
    post/.style =   {->, semithick}
}

\usepackage{amsmath}
\usepackage[utf8]{inputenc} 
\usepackage{algorithmic}
\usepackage{multicol}
\usepackage{tikz,pgfplots,pgfplotstable}
\usepackage{setspace}
\usepackage{indentfirst}
\usepackage{enumitem}
\usepackage{multirow}
\usepackage{colortbl}
\definecolor{pblue}{RGB}{0,0,255}
\definecolor{pwhite}{RGB}{255,255,255}
\definecolor{pblack}{RGB}{0,0,0}
\usepackage{here}
\usepackage{pgfplots}
\usepackage{lipsum}
\usepackage{afterpage}
\usepackage{comment}

\newcommand{\argmax}{\mathop{\mathrm{argmax}}\limits}   

\usetikzlibrary{arrows.meta,chains,decorations.pathreplacing,decorations.pathmorphing}
\usetikzlibrary{shapes,intersections,shapes.geometric, arrows}
\tikzstyle{decision} = [ diamond, draw, fill=blue!20, text width=4.5em, text badly centered, node distance=3cm]  
\tikzstyle{block} = [ rectangle, draw, fill=blue!20, text width=5em, text badly centered, rounded corners, minimum height=4em]  
\tikzstyle{line} = [ draw, -latex]  
\tikzstyle{terminator} = [ draw, ellipse, fill=red!20, node distance=3cm, minimum height=2em]

\begin{document}

\title{A Bayesian Model for Activities Recommendation and Event Structure Optimization Using Visitors Tracking}

\author{Henrique~X.~Goulart,
        Guilherme~A.~Wachs-Lopes}%

\maketitle

\begin{abstract}
In events that are composed by many activities, there is a problem that involves retrieve and management the information of visitors that are visiting the activities. This management is crucial to find some activities that are drawing attention of visitors; identify an ideal positioning for activities; which path is more frequented by visitors. In this work, these features are studied using Complex Network theory. For the beginning, an artificial database was generated to study the mentioned features. Secondly, this work shows a method to optimize the event structure that is better than a random method and a recommendation system that achieves $\sim$95\% of accuracy.
\end{abstract}


\IEEEpeerreviewmaketitle

\section{Introduction}

\IEEEPARstart{I}{n} the last decade, the emergence of new hardware technologies has led many scientists to focus their studies on more complex computational models to solve problems to assist people. One well-known problem category in this area is the Recommendation Systems. Basically, a recommendation system uses statistical techniques to infer people's likes and preferences based on previously known data.

In the last decade, the IoT (Internet of Things) area has gained a lot of space in several industrial branches.  Since its creation in 1999 by Kevin Ashton, it has been possible to manage and track products individually, to checkout automatically in supermarkets, to identify people in homes, to control devices in a large company in an orchestrated way, among many other applications.

IoT is a technology that has as a main focus applications involving people \cite{Ashton2009} and electronic devices in an interactive environment.  In addition, this interaction is expected to be done in the most transparent way possible;  e.g. non-invasive.

There are many applications in the IoT area in order to improve the quality of life \cite{Atzori2010}. These include mobile tiketing applications, which allow users to buy tickets for a particular service from their smartphone, and also view the information about the service provided automatically by the application \cite{broll2009perci}.  Other applications can be found in healthcare, which identifies people as patients and doctor, as well as the objects they will use to be located, which involves automatic data collection and an interconnected environment \cite{vilamovska2009rfid}.

Regarding the health area, there is a great interest in the possibility of sensing patients at a distance, which requires the implementation of wireless sensors integrated into a system that provides information about the patient continuously \cite{niyato2009remote}.

Another branch of IoT is the management of energy consumption in houses;  That is, to propose computational models capable of computing the cost of consumption of each internal equipment based on the information supplied online, and how this consumption can be reduced \cite{buckl2009services}.  Related to the same theme, there are studies that find out ways to optimize the food supply in homes from a set of sensors and a computer system to interconnect everything.  Thus, one can suggest new ways of efficient household supply based on measurement of consumption.

Although the numerous applications that already exist in the area, little is said about the incorporation of this technology for the extraction of statistics in closed events composed of several exhibitions or activities (individual entertainment area).

In these types of events, each person can move differently, visiting different exhibitions and in an order of their own.  At this point, a computational model that could store these information, such as the visitation order for each person present in an event, could contribute for profile analysis of visitors as well as the event itself, such as preferences, behaviors, flows,  best exhibitions, best location of the exhibitions, among many others.  Such a computational model is present in an area that has been gaining attention of researchers in recent years, the Complex Networks Theory.  More generally, this area is based on the union of statistical mathematics and graph theory.

One of the main reasons for the advancement of this area is the evolution of the current hardware, since the algorithms involved have asymptotically high times, in the order of $O(n^3) $, where $n$ is the number of nodes.

In this paper we propose a computational model to analyze the behavior of visitors in closed events. This is done thought two steps: data generation/capture; and data analysis.

\input{Naive}

\section{Proposed Model}
\input{Proposal}
\input{Results}
\input{Conclusion}

\section*{Acknowledgment}

The authors would like to thank Telefônica VIVO partnership with Centro Universitário FEI and IIoT Lab Telef\^{o}nica/FEI (Intelligent Internet of Things Lab) by the funding and hardware resources provided.

\ifCLASSOPTIONcaptionsoff
  \newpage
\fi

\bibliographystyle{IEEEtran}
\bibliography{referencias}

\end{document}

%% file: Naive.tex
\section{Background}

In this section, the concepts that was used in this work will be explained.

\subsection{Genetic Algorithm}
\label{Sec_GACON}

\cite{gendreau2010handbook} affirm, that genetic algorithm is an algorithm based on Neo-Darwinism Theory of evolution; thus, this approach uses concepts like mutation, selection and survival of the strongest individual. Therefore, the author affirms that objective of this algorithm is optimize a solution (hypothesis) based on a fitness function.

\begin{figure}[h]
\centering
\scriptsize
\begin{tikzpicture}
	\draw (0,0)  node (n0) [draw, decision, aspect=2] {Stop condition};
	\draw (0,-3) node (n1) [draw, terminator] {End of process};
	\draw (0,3) node (n2) [rectangle,fill=blue!20,stroke=black,draw] {Generate initial population};
	\draw (0,1.5) node (n3) [block] {Evaluation};
	\draw (-3,-3) node (n4) [block] {Selection};
	\draw (-3,-1.5) node (n5) [block] {Crossover};
	\draw (-3,0) node (n6) [block] {Mutation};
	\draw (-3,1.5) node (n7) [block] {Evaluation};
	\draw (-3,3) node (n8) [block] {Elitism};	
	
	\path[draw, ->] (n0) -> (n1) node [midway, above, sloped] () {Yes};
	\path[draw, ->] (n2) -> (n3);
	\path[draw, ->] (n3) -> (n0);
	\path[draw, ->] (n0) -> (n4) node [midway, above, sloped] () {No};
	\path[draw, ->] (n4) -> (n5);
	\path[draw, ->] (n5) -> (n6);
	\path[draw, ->] (n6) -> (n7);
	\path[draw, ->] (n7) -> (n8);
	\path[draw, ->] (n8) -> (n0);
	
\end{tikzpicture}
\caption{Operation of a genetic algorithm.}
\label{Fig_AG}
\end{figure}
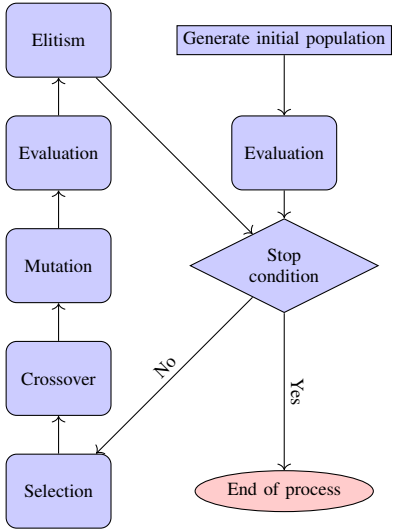

A typical genetic algorithm operates following the steps represented in Figure \ref{Fig_AG} \cite{mitchell1997machine}.

\subsection{Naive Bayes Model}
\label{Sec_NB}

The Naive Bayes model is a bayesian network model and n-gram model very used in the machine learning area \cite{russell2003}. This model is a probabilistic model that correlates variables using fewer computer resources than other n-gram models because these other models require a joint probability table, that table grows exponentially with the addition of new variables.

In this model the variables are separated into \emph{causes} and \emph{effects}; thus, it is proposed that the \emph{effects} are conditionally independents each other, as from \emph{causes}. That is represented in the Equation (\ref{Eq_CECompl}).

\begin{equation}
P(e|c_1, ...,c_n) = \frac {P(e) \prod_{i=1}^{n} P(c_i|e)}{\gamma}
\label{Eq_CECompl}
\end{equation}

Therefore, this model not requires a joint probability table to store the probabilities. The $\gamma$ is a constant factor for all \emph{effects}, thus, it can be removed from inference, resulting in Equation (\ref{Eq_CECompl2}).

\begin{equation}
\varphi = \argmax_{k} (P(e_k) \prod_{i=1}^{n} P(c_i|e_k))
\label{Eq_CECompl2}
\end{equation}

%% file: Proposal.tex
The proposed model for this work consists of extracting information from an event (a group of activities and visitors) and using that information to improve visitor experience. The information of visits pattern can be collected once the event is modeled as Figure \ref{Fig_GrafoGeral} illustrates.

\subsection{Datasets}
\label{Sec_dataset}

For this work, an artificial graph was used to create visits patterns. 

\begin{figure}[h]
\centering
\begin{tikzpicture}[scale=0.75]

	\draw (-3,0) node (n1) [circle,fill=blue!20,stroke=black,draw]{Activity 3};
	\draw (0,0) node (n2) [circle,fill=green!20,stroke=black,draw]{Activity 2};	
	\draw (3,0) node (n3) [circle,fill=green!20,stroke=black,draw]{Activity 5};
	\draw (6,0) node (n4) [circle,fill=green!20,stroke=black,draw]{Activity 1};
	
	\path[draw,->, -triangle 90] (n1) -> (n2);
	\path[draw,->, -triangle 90] (n2) -> (n3);
	\path[draw,->, -triangle 90] (n3) -> (n4);

\end{tikzpicture}
\caption{Visitor pattern.}
\label{Fig_VisPatt}
\end{figure}
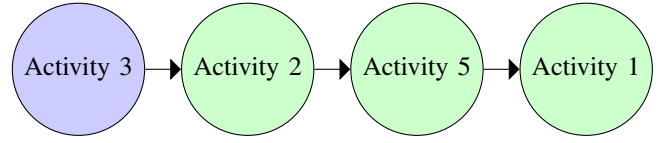

Initially, each visitor pattern (visit pattern of a visitor) has only 1 visit (initial visit), so that all activities have the same number of visitors at the beginning. That visitor pattern is modified according to the noise factor, which determines how many random activities will be added (that do not exist) to the pattern, this is represented in the Figure \ref{Fig_VisPatt}.

\subsection{Training}

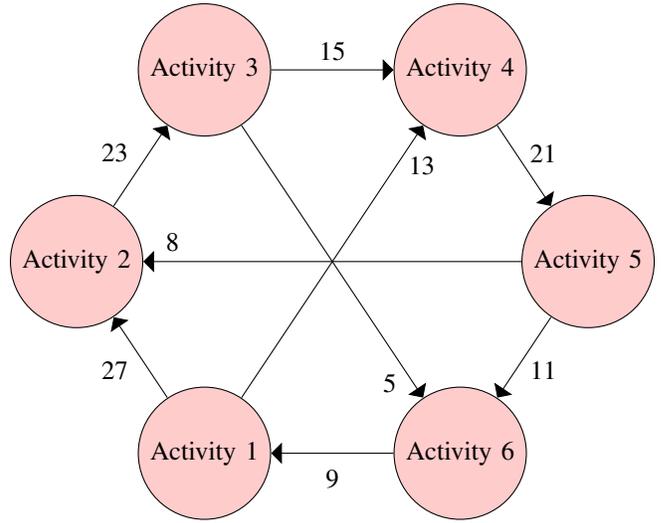
\begin{figure}[h]
\centering
\begin{tikzpicture}[scale=0.85]

	\draw (0,0) node (n1) [circle,fill=red!20,stroke=black,draw]{Activity 1};
	\draw (-2,3) node (n2) [circle,fill=red!20,stroke=black,draw]{Activity 2};	
	\draw (0,6) node (n3) [circle,fill=red!20,stroke=black,draw]{Activity 3};
	\draw (4,6) node (n4) [circle,fill=red!20,stroke=black,draw]{Activity 4};
	\draw (6,3) node (n5) [circle,fill=red!20,stroke=black,draw]{Activity 5};
	\draw (4,0) node (n6) [circle,fill=red!20,stroke=black,draw]{Activity 6};
	
	\path[draw,->, -triangle 90] (n1) -> (n2);
	\path[draw,->, -triangle 90] (n2) -> (n3);
	\path[draw,->, -triangle 90] (n3) -> (n4);
	\path[draw,->, -triangle 90] (n4) -> (n5);
	\path[draw,->, -triangle 90] (n5) -> (n6);
	\path[draw,->, -triangle 90] (n6) -> (n1);
	\path[draw,->, -triangle 90] (n1) -> (n4);
	\path[draw,->, -triangle 90] (n3) -> (n6);
	\path[draw,->, -triangle 90] (n5) -> (n2);

	\node (a) at (2,6.3) {15};
	\node (b) at (-1.4,4.7) {23};
	\node (c) at (-1.4,1.3) {27};
	\node (d) at (2,-0.4) {9};
	\node (e) at (5.3,4.7) {21};
	\node (f) at (5.3,1.3) {11};
	\node (g) at (-0.5 ,3.3) {8};
	\node (h) at (2.9,1.1) {5};
	\node (i) at (3.4,4.5) {13};

\end{tikzpicture}
\caption{Graph of an event.}
\label{Fig_GrafoGeral}
\end{figure}

To create the model represented in Figure \ref{Fig_GrafoGeral}, the graph is generated from a junction of all visitors pattern, so that each edge in the graph is weighted according to the number of visitors which visits the activities connected to this edge.

For computer model representation, this model should be interpreted as a matrix. The matrix $W$ (\ref{Eq_MatrizPeso}) represents the graph of Figure \ref{Fig_GrafoGeral}.

\begin{equation}
W=\begin{pmatrix}      
0&27&0&13&0&0 \\      
0&0&23&0&0&0 \\ 
0&0&0&15&0&5 \\   
0&0&0&0&21&0 \\
0&8&0&0&0&11 \\
9&0&0&0&0&0 \\ 
\end{pmatrix}
\label{Eq_MatrizPeso}
\end{equation}

The Naive Bayes model needs a probabilistic value, then the matrix $W$ should be transformed into a probabilistic matrix $P$ so that in each position it is applied the Equation (\ref{Eq_Norm}).

\begin{equation}
W_{i,j} = \frac{W_{i,j}}{\sum_{i} W_{i,j}}
\label{Eq_Norm}
\end{equation}

Each position on matrix $W$ is incremented by 1, because, if the result is 0, it will be a problem in posterior calculations; the term $sumW_j$ is the sum of all positions of the column $j$ in matrix $W$ (with the positions already incremented). The matrix $P$ is represented below:

\begin{equation}
P=\begin{pmatrix}      
0.067&0.7&0.034&0.412&0.037&0.045 \\      
0.067&0.025&0.828&0.09&0.037&0.045 \\ 
0.067&0.025&0.034&0.471&0.037&0.273 \\   
0.067&0.025&0.034&0.09&0.815&0.045 \\
0.067&0.225&0.034&0.09&0.037&0.545 \\
0.67&0.025&0.034&0.09&0.037&0.045 \\ 
\end{pmatrix}
\label{Eq_MatrizProb}
\end{equation}

\subsection{Genetic Algorithm}

As stated in Section \ref{Sec_GACON}, genetic algorithm (G.A) is an algorithm used to optimize some solution to a problem. This work proposes to improve visitor experience, which implies to improve the activities positions in an event using the G.A (using activities positions $(x,y)$ as G.A chromosome).

\begin{figure}[h]
\centering
  	\begin{tikzpicture}[
       start chain = going right,
     node distance = 0pt,
MyStyle/.style={draw, minimum width=2em, minimum height=2em,outer sep=0pt, on chain},]

		\node [MyStyle,fill=blue!20] (1) {$(5,0)$};
		\node [MyStyle,fill=blue!20] (2) {$(6,3)$};
		\node [MyStyle,fill=blue!20] (3) {$(1,3)$};
		\node [MyStyle,fill=blue!20] (4) {$(3,7)$};
		\node [MyStyle,fill=blue!20] (5) {$(0,2)$};

	\end{tikzpicture}
	
	\begin{tikzpicture}[
       start chain = going right,
     node distance = 0pt,
MyStyle/.style={draw, minimum width=2em, minimum height=2em,outer sep=0pt, on chain},]

		\node [MyStyle,fill=red!20] (1) {$(6,6)$};
		\node [MyStyle,fill=red!20] (2) {$(3,4)$};
		\node [MyStyle,fill=red!20] (3) {$(5,0)$};
		\node [MyStyle,fill=red!20] (4) {$(5,8)$};
		\node [MyStyle,fill=red!20] (5) {$(9,0)$};

	\end{tikzpicture}
	
	\begin{tikzpicture}[
  		>=stealth',
  		pos=.8,
  		photon/.style={decorate,decoration={snake,post length=1mm}}
	]
  		\draw[->] (0,0) -- node[above right] {} (-90:1);
	\end{tikzpicture}	
	
	\begin{tikzpicture}[
       start chain = going right,
     node distance = 0pt,
MyStyle/.style={draw, minimum width=2em, minimum height=2em,outer sep=0pt, on chain},]

		\node [MyStyle,fill=black!100] (1) {\color{white}(5,0)};
		\node [MyStyle,fill=red!20] (2) {$(3,4)$};
		\node [MyStyle,fill=black!100] (3) {\color{white}(5,0)};
		\node [MyStyle,fill=blue!20] (4) {$(3,7)$};
		\node [MyStyle,fill=blue!20] (5) {$(0,2)$};

	\end{tikzpicture}
	
	\begin{tikzpicture}[
  		>=stealth',
  		pos=.8,
  		photon/.style={decorate,decoration={snake,post length=1mm}}
	]
  		\draw[->] (0,0) -- node[above right] {} (-90:1);
	\end{tikzpicture}	
	
	\begin{tikzpicture}[
       start chain = going right,
     node distance = 0pt,
MyStyle/.style={draw, minimum width=2em, minimum height=2em,outer sep=0pt, on chain},]

		\node [MyStyle,fill=blue!20] (1) {$(5,0)$};
		\node [MyStyle,fill=red!20] (2) {$(3,4)$};
		\node [MyStyle,fill=red!20] (3) {$(6,6)$};
		\node [MyStyle,fill=blue!20] (4) {$(3,7)$};
		\node [MyStyle,fill=blue!20] (5) {$(0,2)$};

	\end{tikzpicture}

\caption{Conflict in crossover and their treatment.}
\label{Fig_Conflito}
\end{figure}
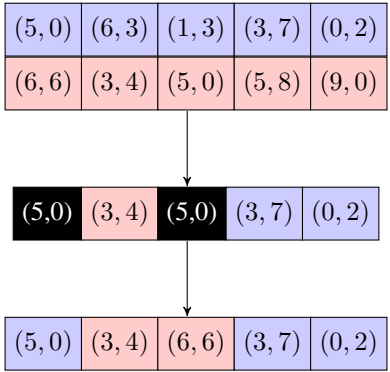

In crossover step, it was used a 2-point ordered crossover, but as observed in Figure \ref{Fig_Conflito} a usual approach can produce some conflicts, to solve those conflicts it is necessary to replace the duplicated position by the first position that not used in the individual. Moreover, the points used in crossover cannot be the first one of individual, thus the first position is fixed then the G.A must find a solution on this position.

To mutate the individuals, it was applied two types of mutation with the same chance. thus, the maximum probability for mutation is 50\%. One of them swap two positions of individual; and the second, swap one position of the individual with a position not used by individual.

To evaluate each individual the Equation (\ref{Eq_Avaliacao}) was applied. Therefore, it is possible to consider the frequency of visitors in the activity (that information is stored in the edges of graph stated before) and their position regarding others positions.

\begin{equation}
Fitness = \sum\limits_{(n_i,n_j) \in N^2} \frac{W(n_i,n_j)}{d(i,j)}
\label{Eq_Avaliacao}
\end{equation}

\subsection{Recommendation}

As stated in Section \ref{Sec_NB}, to create a recommendation system based on Naive Bayes, it is necessary to subdivide the variables of this work into \emph{causes} and \emph{effects}.

Using the matrix $W$ is possible to separate your lines as \emph{previous activities} and the columns as \emph{next activities}. Thus, the \emph{previous activities} is interpreted as \emph{causes} and \emph{next activities} as \emph{effects}.

Thereby, is possible to establish probabilistically, the best activity to recommend using as \emph{history} ($n$-last) the activities already visited. The Equation (\ref{Eq_AvaliacaoNB}) describe a search of best (\emph{argmax}) activity $k$.

\begin{equation}
\varphi = \argmax_{k} (P(Next_k) \prod_{i=1}^{n} P(Previous_i|Next_k))
\label{Eq_AvaliacaoNB}
\end{equation}

The term $P(Next_k)$ is the \emph{a priori} probability of activity $k$ to be recommended; numerically, is a number of visitors that visit the activity $k$ normalized by the total of visits in the event. The term $P(Previous_i|Next_k)$ is the probability of recommending the activity $k$, given that the activity $i$ was visited; numerically, is the value of the position $(i,k)$ in matrix W.

In order to avoid precision loss in the calculations in Equation (\ref{Eq_AvaliacaoNB}), was used a technique which consists of conveying to \emph{log}. That conveying is described below:

\indent $P(Next_k) \prod_{i=1}^{n} P(Previous_i|Next_k)$ \\
\indent $ = log_{2}(P(Next_k)) + \sum_{i=1}^{n} log_{2}(P(Previous_i|Next_k))$

In this work, it was considered that the last visit has twice the weight compared to the previous.

For modeling that effect, a coefficient $\alpha$ is used on \emph{a posteriori} terms according to Equation (\ref{Eq_NormAlpha}).

\begin{equation}
R = log_{2}(P(Next_k)) + \sum_{i=1}^{n} \alpha_i log_{2}(P(Previous_i|Next_k))
\label{Eq_NormAlpha}
\end{equation}

Thus, the sum guarantees the precision of calculation so that come back to the probabilistic value exponentiating the result $R$: $2^R$.

%% file: Results.tex
\section{Results}

In this section, results of experiments applied to the genetic algorithm (G.A) will be showed.

\subsection{Mutation and Crossover}

An experiment was proposed to understand the behavior of G.A in function of crossover chance and mutation parameters. To perform this experiment, it was used a graph (represented in Figure \ref{Fig_GrafoGeral}) composed by 100 activities and 470000 visits in total; and the G.A optimized the solution over 5000 generations. 

\begin{figure}[h]
\centering
\includegraphics[width=0.54\textwidth]{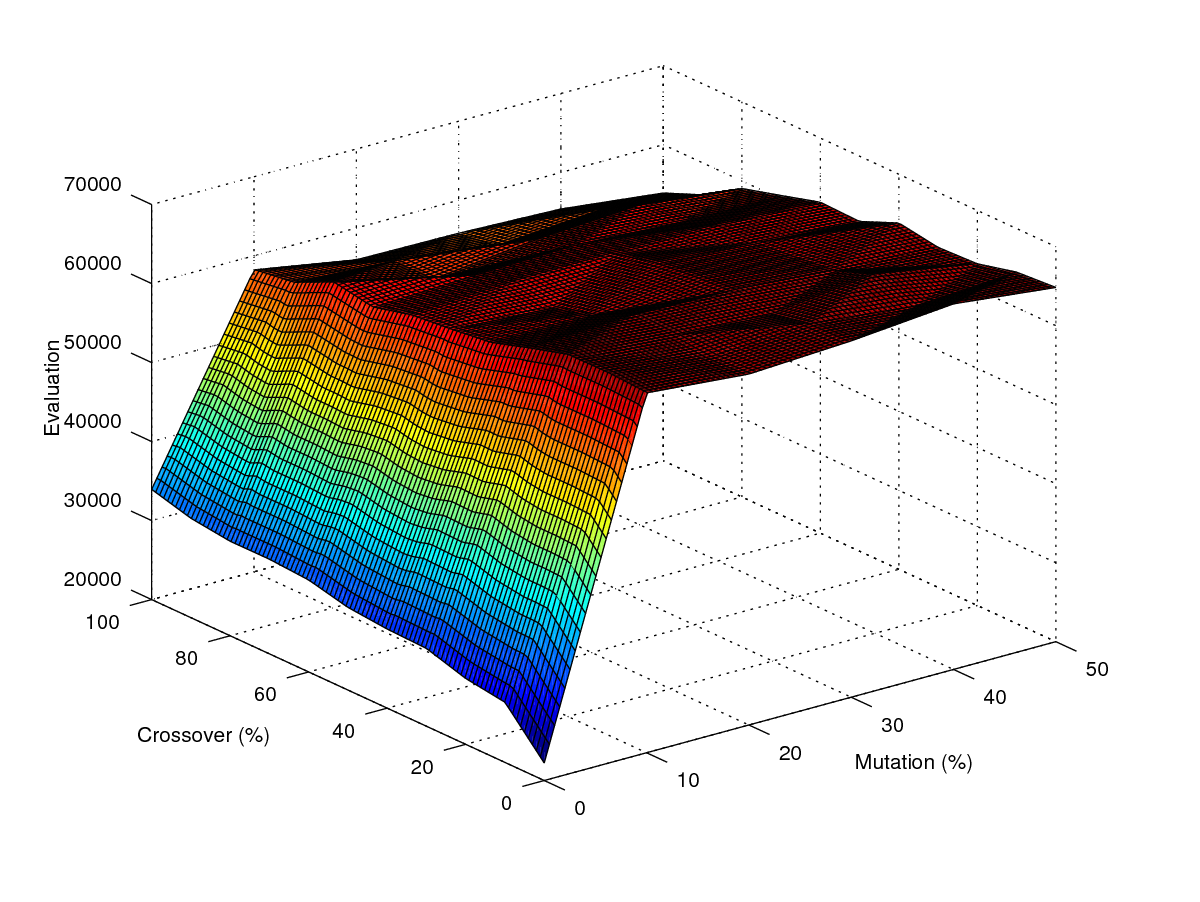}
\caption{Illustration of the G.A performance in function of mutation and crossover values.}
\label{Fig_Mutacao}
\end{figure}

As illustrated in Figure \ref{Fig_Mutacao}, both parameters is important to evaluation. Therefore, using crossover chance at 10\% and mutation at 40\%, the G.A performed best results.

\subsection{Genetic Algorithm vs Random Algorithm}

\begin{figure}[h]
\centering
\includegraphics[width=0.5\textwidth]{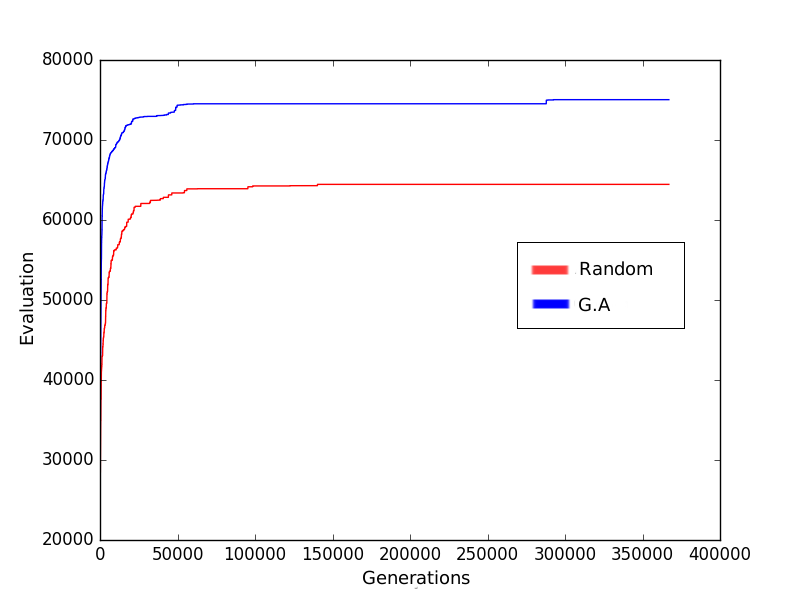}
\caption{Illustration of performance of G.A compared to Random Algorithm.}
\label{Fig_NotasBNC}
\end{figure}

In order to compare the performance of G.A with other methods as illustrated in Figure \ref{Fig_NotasBNC}, the Random Algorithm was used as baseline. This method, use an initial individual of G.A and each generation two positions of individual is swapped.

\subsection{Recommendation}

\begin{figure}[h]
\scalebox{.77}{
\centering

\begin{tabular}{@{}cccc@{}}

\begin{tikzpicture}[scale=0.8]

	\draw (-3,0) node (n1) [circle,fill=blue!20,stroke=black,draw]{Activity 7};
	\draw (0,0) node (n2) [circle,fill=blue!20,stroke=black,draw]{Activity 2};	
	\draw (3,0) node (n3) [circle,fill=blue!20,stroke=black,draw]{Activity 3};
	\draw (6,0) node (n4) [circle,fill=blue!20,stroke=black,draw]{Activity 1};
	\draw (9,0) node (n5) [circle,fill=blue!20,stroke=black,draw]{Activity 5};
	
	\path[draw,->, -triangle 90] (n1) -> (n2);
	\path[draw,->, -triangle 90] (n2) -> (n3);
	\path[draw,->, -triangle 90] (n3) -> (n4);
	\path[draw,->, -triangle 90] (n4) -> (n5);

\end{tikzpicture}

\\

\begin{tabular}{l}
$\Downarrow$
\end{tabular}

\\

\begin{tikzpicture}[scale=0.8]

	\draw (-3,0) node (n1) [circle,fill=yellow!20,stroke=black,draw]{Activity 7};
	\draw (0,0) node (n2) [circle,fill=yellow!20,stroke=black,draw]{Activity 2};	
	\draw (3,0) node (n3) [circle,fill=yellow!20,stroke=black,draw]{Activity 3};
	\draw (6,0) node (n4) [circle,fill=cyan!60,stroke=black,draw,dashed,minimum size=1.8cm]{};
	\draw (9,0) node (n5) [circle,fill=cyan!60,stroke=black,draw,dashed,minimum size=1.8cm]{};
	
	\path[draw,->, -triangle 90] (n1) -> (n2);
	\path[draw,->, -triangle 90] (n2) -> (n3);
	\path[draw,->, -triangle 90] (n3) -> (n4);
	\path[draw,->, -triangle 90] (n4) -> (n5);

\end{tikzpicture}

\\

\begin{tikzpicture}[scale=0.8]

	\draw (-3,0) node (n1) [circle,fill=yellow!20,stroke=black,draw]{Activity 7};
	\draw (0,0) node (n2) [circle,fill=yellow!20,stroke=black,draw]{Activity 2};	
	\draw (3,0) node (n3) [circle,fill=yellow!20,stroke=black,draw]{Activity 3};
	\draw (6,0) node (n4) [circle,fill=green!20,stroke=black,draw]{Activity 4};
	\draw (9,0) node (n5) [circle,fill=cyan!60,stroke=black,draw,dashed,minimum size=1.8cm]{};
	
	\path[draw,->, -triangle 90] (n1) -> (n2);
	\path[draw,->, -triangle 90] (n2) -> (n3);
	\path[draw,->, -triangle 90] (n3) -> (n4);
	\path[draw,->, -triangle 90] (n4) -> (n5);

\end{tikzpicture}

\\

\begin{tikzpicture}[scale=0.8]

	\draw (-3,0) node (n1) [circle,fill=black!40,stroke=black,draw]{Activity 7};
	\draw (0,0) node (n2) [circle,fill=yellow!20,stroke=black,draw]{Activity 2};	
	\draw (3,0) node (n3) [circle,fill=yellow!20,stroke=black,draw]{Activity 3};
	\draw (6,0) node (n4) [circle,fill=yellow!20,stroke=black,draw]{Activity 4};
	\draw (9,0) node (n5) [circle,fill=green!20,stroke=black,draw]{Activity 5};
	
	\path[draw,->, -triangle 90] (n1) -> (n2);
	\path[draw,->, -triangle 90] (n2) -> (n3);
	\path[draw,->, -triangle 90] (n3) -> (n4);
	\path[draw,->, -triangle 90] (n4) -> (n5);

\end{tikzpicture}

\end{tabular}

}
\caption{Experiment to test the recommendation system using 3 previous visits as history and cutting 2 visits.}
\label{Fig_VisPattTest}
\end{figure}
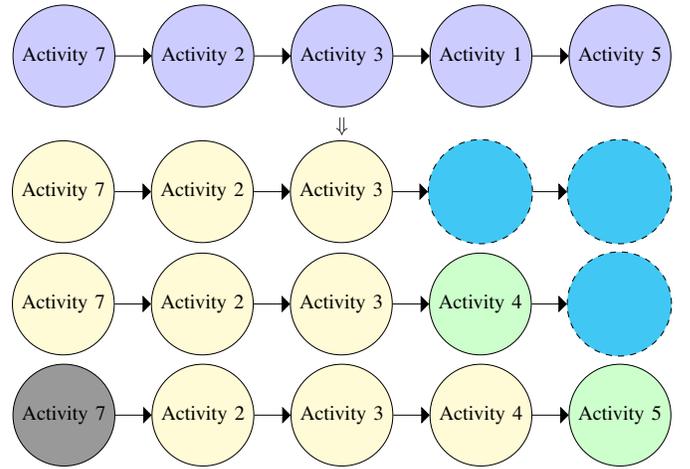

To perform a validation experiment of the recommendation system proposed, some visits of visitors patterns was removed (cut). Then, the system will recommend visits to reconstruct the visitor, based on previous visits in pattern (history). This experiment is illustrated in Figure \ref{Fig_VisPattTest}, the visit colors represents the following: blue as original pattern, yellow as the history, cyan as the activity to recommend, green as the recommendation and gray as a visit not used.

To evaluate the recommendations, it was applied the Equation (\ref{Eq_PV}) in order to compare only the pattern recommended and original ones. Therefore, $a$ is the visit index (e.g. \emph{index of} Activity 5 is 5) of original pattern and $b$ is the recommended ones.

\begin{figure}[h]
\scalebox{.77}{
\centering

\begin{tabular}{@{}cc@{}}

\begin{tikzpicture}[scale=0.8]

	\draw (-3,0) node (n1) [circle,fill=black!40,stroke=black,draw]{Activity 7};
	\draw (0,0) node (n2) [circle,fill=black!40,stroke=black,draw]{Activity 2};	
	\draw (3,0) node (n3) [circle,fill=black!40,stroke=black,draw]{Activity 3};
	\draw (6,0) node (n4) [circle,fill=blue!20,stroke=black,draw]{Activity 1};
	\draw (9,0) node (n5) [circle,fill=blue!20,stroke=black,draw]{Activity 5};
	
	\path[draw,->, -triangle 90] (n1) -> (n2);
	\path[draw,->, -triangle 90] (n2) -> (n3);
	\path[draw,->, -triangle 90] (n3) -> (n4);
	\path[draw,->, -triangle 90] (n4) -> (n5);

\end{tikzpicture}
\\
\begin{tikzpicture}[scale=0.8]

	\draw (-3,0) node (n1) [circle,fill=black!40,stroke=black,draw]{Activity 7};
	\draw (0,0) node (n2) [circle,fill=black!40,stroke=black,draw]{Activity 2};	
	\draw (3,0) node (n3) [circle,fill=black!40,stroke=black,draw]{Activity 3};
	\draw (6,0) node (n4) [circle,fill=green!20,stroke=black,draw]{Activity 4};
	\draw (9,0) node (n5) [circle,fill=green!20,stroke=black,draw]{Activity 5};
	
	\path[draw,->, -triangle 90] (n1) -> (n2);
	\path[draw,->, -triangle 90] (n2) -> (n3);
	\path[draw,->, -triangle 90] (n3) -> (n4);
	\path[draw,->, -triangle 90] (n4) -> (n5);

\end{tikzpicture}

\end{tabular}

}
\caption{Visualization of original pattern and the recommended visits.}
\label{Fig_pattcomp}
\end{figure}
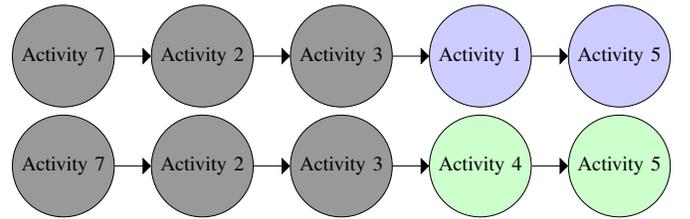

The Figure \ref{Fig_pattcomp} shows which visits will be compared following the previous example, where blue visits are the original, green are the recommended and grey are the visits not used in evaluation.

\begin{equation}
Evaluation~=~\frac{\sum_{i=1}^{x} a_{i}b_{i}}{\sum_{i=1}^{x} min(a_{i},b_{i})^2}
\label{Eq_PV}
\end{equation}

The test is proposed consists to variate the cut parameter and the number of previous visits considered. Thus, is possible to verify how the system works in different scenarios.

\begin{figure}[h]
\centering
\includegraphics[width=0.5\textwidth]{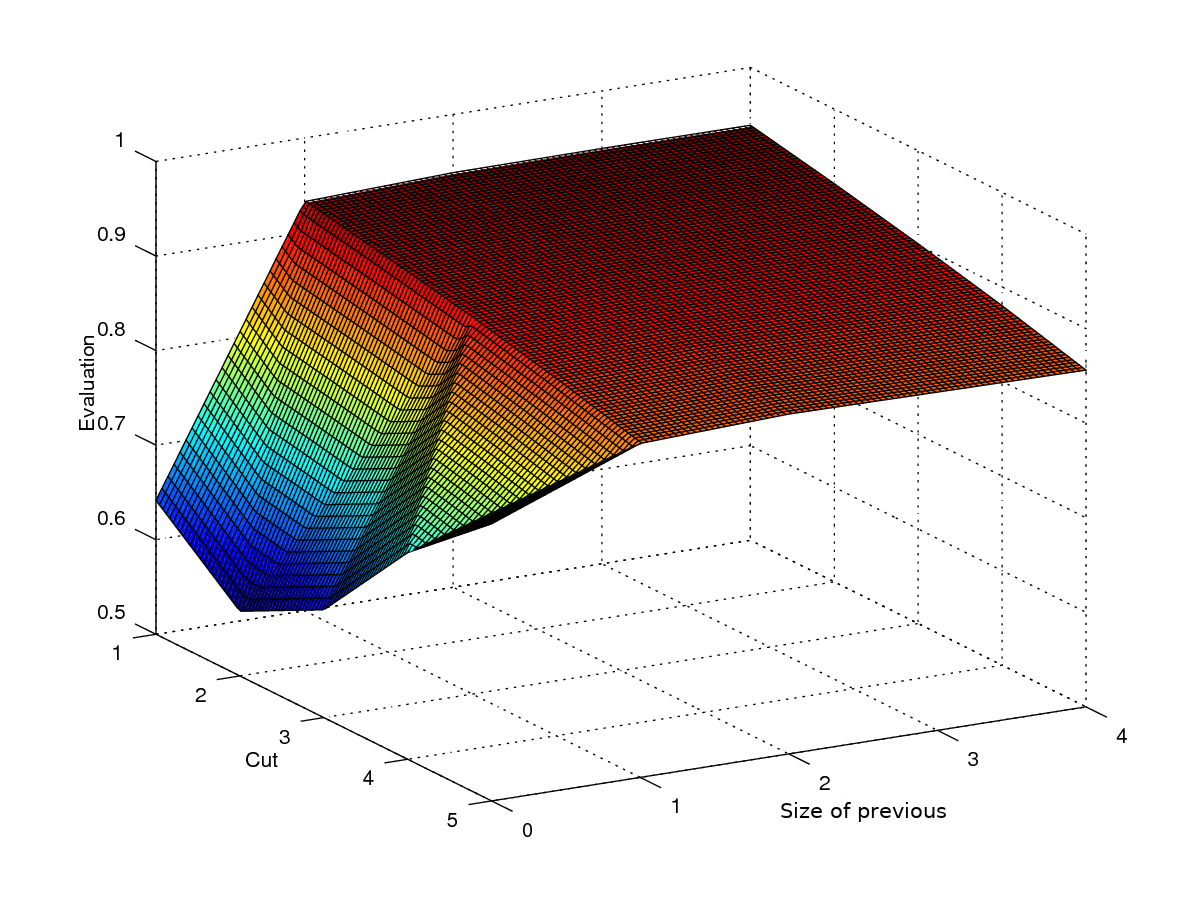}
\caption{Illustration of recommendation system behavior in different scenarios.}
\label{Fig_Recom3Dpart}
\end{figure}

The Figure \ref{Fig_Recom3Dpart} represents the precision of the recommendation system in different scenarios (i.e. different values for $cut$ and \emph{size of previous} parameters). In this experiment, it was used 20000 visitor patterns and each evaluation is a mean of all these 20000 evaluations. As observed, except for the size of previous equal to zero, all other scenarios have similar results; and the size of previous implies that error of one recommendation interferes with the next, causing a grow of error in inference.

%% file: Conclusion.tex
\section{Conclusion}

To study the visitors behavior in events, it was proposed a model graph-based to optimize the event structure that considers all visitors patterns; that optimization also considers the geographic location of activities. Furthermore, using that same patterns, it was proposed a recommendation system that suggests the next activity to visit based on previous visits.

The genetic algorithm (G.A) was used to optimize the event structure. To validate results, the random algorithm was applied to compare results. As observed, it is visible that the G.A is better than random algorithm.

The recommendation system was developed using the Naive Bayes and producing promising results, because it achieves $\sim$95\% of accuracy in some scenarios.

For future work, a real database can be applied to compare and validate the results with the artificial database. Using another model with G.A to create a hybrid model, can produce better results.

Using another model also with Naive Bayes can improve the accuracy the recommendation system, maybe a model that consider the semantic of words like LSA, because that model not uses a large database for training.

Finally, the proposed model reached the objectives and can be improved to achieve better results.